\let\oldyear\year
\documentclass{ieeeaccess}

\let\year\oldyear
\NewSpotColorSpace{PANTONE}
\AddSpotColor{PANTONE} {PANTONE3015C} {PANTONE\SpotSpace 3015\SpotSpace C} {1 0.3 0 0.2}
\SetPageColorSpace{PANTONE}
\definecolor{accessblue}{cmyk}{1, 0.3, 0, 0.2}
\definecolor{greycolor}{cmyk}{0,0,0,.8}
\usepackage{amsmath,amsfonts}
\usepackage{algpseudocode}
\usepackage{algorithm}
\usepackage{array}
\usepackage{csvsimple}
\usepackage[caption=false,font=normalsize,labelfont=sf,textfont=sf]{subfig}
\usepackage[colorlinks=true,linkcolor=blue]{hyperref}%
\usepackage{url}
\usepackage{graphicx}
\usepackage{datatool}
\usepackage{pgfplots}
\usetikzlibrary{matrix,positioning}
\pgfplotsset{compat=newest} 
\usepgfplotslibrary{groupplots,fillbetween}
\usepackage{tikzscale}

\def\BibTeX{{\rm B\kern-.05em{\sc i\kern-.025em b}\kern-.08em
    T\kern-.1667em\lower.7ex\hbox{E}\kern-.125emX}}

\algnewcommand\algorithmicinput{\textbf{Input:}}
\algnewcommand\algorithmicoutput{\textbf{Output:}}
\algnewcommand\Input{\item[\algorithmicinput]}%
\algnewcommand\Output{\item[\algorithmicoutput]}%

\begin{document}
\history{Date of acceptance 6 February 2024, date of current version 10 February 2024.}
\doi{10.1109/ACCESS.2024.3365517}

\title{Adversarial Robustness on Image Classification with $k$-means}
\author{\uppercase{Rollin Omari}\authorrefmark{1},
\uppercase{Junae Kim}\authorrefmark{1}, and 
\uppercase{Paul Montague}\authorrefmark{1}}

\address[1]{Defence Science and Technology Group, Australian 
Department of Defence, Edinburgh, SA, 5111 Australia.}

\markboth
{Omari \headeretal: Adversarial Robustness on Image Classification with $k$-means}
{Omari \headeretal: Adversarial Robustness on Image Classification with $k$-means}

\corresp{Corresponding author: Rollin Omari (e-mail: rollin.omari@defence.gov.au).}

\begin{abstract}
Attacks and defences in adversarial machine 
learning literature have primarily focused on 
supervised learning. However, it remains an open 
question whether existing methods and strategies 
can be adapted to unsupervised learning approaches. 
In this paper we explore the challenges and strategies 
in attacking a $k$-means clustering algorithm and in 
enhancing its robustness against adversarial 
manipulations. We evaluate the vulnerability of 
clustering algorithms to adversarial attacks on 
two datasets (MNIST and Fashion-MNIST), emphasising 
the associated security risks. Our study investigates 
the impact of incremental attack strength on training, 
introduces the concept of transferability between 
supervised and unsupervised models, and highlights 
the sensitivity of unsupervised models to sample 
distributions. We additionally introduce and evaluate 
an adversarial training method that improves testing 
performance in adversarial scenarios, and we highlight 
the importance of various parameters in the proposed 
training method, such as continuous learning, centroid 
initialisation, and adversarial step-count. Overall, 
our study emphasises the vulnerability of unsupervised 
learning and clustering algorithms to adversarial 
attacks and provides insights into potential defence 
mechanisms. 
\end{abstract}


\titlepgskip=-21pt

\maketitle

\section{Introduction}\label{intro}

In recent years, the field of machine learning has 
witnessed remarkable progress, with advancements 
particularly in unsupervised learning techniques, 
providing solutions to complex problems where unlabelled 
data is plentiful. However, this progress is now 
accompanied by a growing concern for reliability 
and adversarial robustness \cite{xu2020adversarial,
chakraborty2021survey}. As unsupervised learning 
becomes integral to various artificial intelligence 
applications, its robustness becomes synonymous with 
the reliability of the entire system. A failure to 
address adversarial vulnerabilities may lead to undesirable 
consequences, ranging from biased decision-making to 
compromised security \cite{martins2020adversarial}. 
Hence, understanding and addressing adversarial 
vulnerabilities in unsupervised learning is crucial, 
as it directly impacts the real-world applicability 
of such models \cite{chhabra2020suspicion}. 

Amongst the unsupervised techniques, clustering is 
potentially the most popular. The primary objective 
of clustering is to partition data such that similar 
samples are grouped together, while dissimilar ones 
are kept in separate clusters \cite{dubes1980clustering}. 
Machine learning literature contains a broad range 
of clustering algorithms and their applications, 
including but not limited to density-based (e.g., 
DBSCAN \cite{khan2014dbscan}), distribution-based 
(e.g., Gaussian mixture model \cite{reynolds2009gaussian}), 
centroid-based (e.g., $k$-means \cite{hartigan1979algorithm}) 
and hierarchical-based (e.g., BIRCH \cite{zhang1997birch}) 
clustering algorithms.

The $k$-means clustering algorithm in particular, 
iteratively assigns each sample to the cluster with 
the closest center, relying on similarity measurements 
to update the cluster centres. Due to its simplicity 
and versatility, $k$-means is often used in the initial 
stages of data exploration and analysis. However, these 
traits also makes it highly vulnerable to adversarial 
attacks \cite{chhabra2020suspicion}. Namely, $k$-means 
relies on the Euclidean distance between samples and 
cluster centres to assign samples to clusters. Hence 
perturbed samples can disrupt the clustering process 
by pushing samples across cluster boundaries, leading 
to different clustering results \cite{skillicorn2009adversarial}.

\begin{figure}[!t]
\centering
\subfloat[]{\includegraphics[width=0.2\textwidth]
{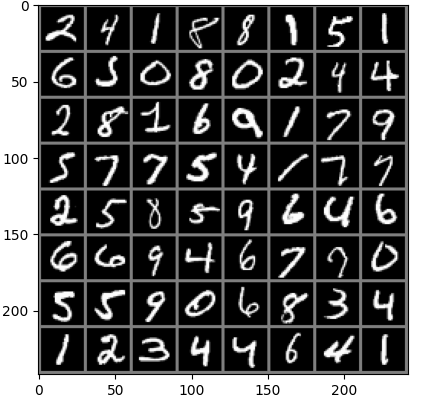}}\hspace*{2em}
\subfloat[]{\includegraphics[width=0.2\textwidth]
{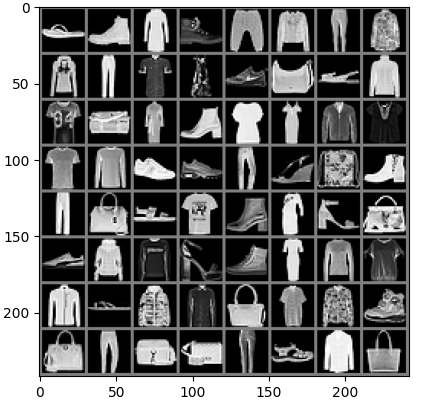}}\\
\subfloat[]{\includegraphics[width=0.2\textwidth]
{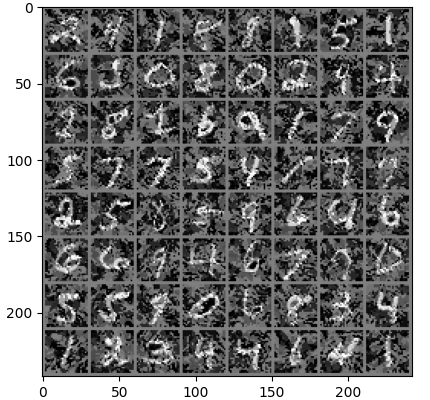}}\hspace*{2em}
\subfloat[]{\includegraphics[width=0.2\textwidth]
{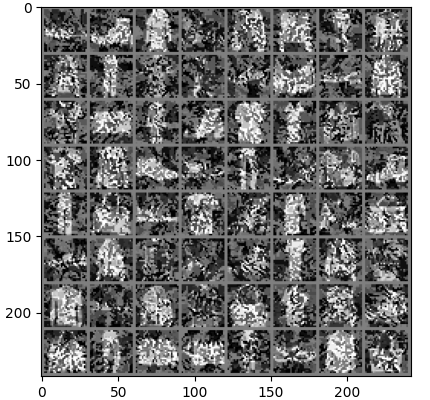}}
\caption{Collages of $8 \times 8$ randomly-picked 
images of handwritten digits and fashion items, 
respectively from the MNIST and Fashion-MNIST 
training datasets. Both datasets have a total 
of 70,000 samples with 60,000 images for training 
and 10,000 for testing. Collages (a) and (b) 
contain clean examples, while collages (c) and 
(d) contain adversarial examples of (a) and (b). 
For both (c) and (d), the adversarial examples 
are generated with I-FGSM.}
\label{collages}
\end{figure}

Biggio et al. \cite{biggio2013data} were one of the 
first to consider adversarial attacks to clustering, 
where they described the obfuscation and poisoning 
attack settings, and provided results on single-linkage 
hierarchical clustering. They also considered evasion 
attacks against surrogate models in a limited-knowledge 
scenario \cite{biggio2013evasion}. Recently, Crussell 
and Kegelmeyer \cite{crussell2015attacking} proposed 
a poisoning attack specific to DBSCAN clustering, and 
Chhabra et al. \cite{chhabra2020suspicion} proposed 
a black-box attack for $k$-means clustering on a subset 
of the MNIST dataset, while Demontis et al. \cite{demontis2019adversarial} 
provided a comprehensive evaluation on transferability 
and the factors contributing to the somewhat model-agnostic 
nature of adversarial examples.

To mitigate the threat of adversarial examples, 
a large variety of defence methods have been proposed, 
including adversarial training, which involves 
incorporating adversarial examples during the training 
process \cite{goodfellow2014explaining, madry2017towards, 
zhang2019theoretically}; input transformations, which 
involve altering the input data via augmentation, 
smoothing or normalisation to improve model robustness 
\cite{xie2017mitigating, guo2017countering}; de-noising, 
which removes or reduces noise from input data with 
filtering techniques \cite{mustafa2019image}; and 
certified defence, which provides bounds and guarantees 
for a model's output \cite{cohen2019certified, 
raghunathan2018certified}. However, these existing 
methods are heavily specialised towards supervised 
training. Among them, TRadeoff-inspired Adversarial 
DEfense via Surrogate-loss minimization (TRADES) 
\cite{zhang2019theoretically} and Projected Gradient 
Descent Adversarial Training \cite{madry2017towards} 
are the most popular adversarial training methods. as 
they provide consistent improvements on robustness 
against various attacks.

In this paper, we take inspiration from these methods 
and introduce an adversarial training algorithm designed 
to enhance the robustness of a $k$-means clustering 
algorithm. Our method involves manipulating proportions 
of clean and perturbed samples in training data and 
iteratively training $k$-means in a continuous manner. 
An underlying intention of this method is to establish 
a much needed baseline for adversarial training for 
unsupervised algorithms. Our experimental results, 
conducted on widely recognised benchmark datasets, 
(i.e., MNIST \cite{lecun1998mnist} and Fashion-MNIST 
\cite{xiao2017fashion}, see Fig. \ref{collages} for 
examples from each set), demonstrate the effectiveness 
of our simple adversarial learning algorithm. It 
significantly enhances the robustness of the clustering 
algorithm while also maintaining its overall performance. 
Importantly, since our method is directed towards 
manipulating training data distributions, it can be 
seamlessly integrated into various unsupervised 
learning frameworks to bolster their robustness.

In summary, the key contributions of this paper are 
threefold. Firstly, we introduce an \emph{unsupervised} 
adversarial training method and demonstrate its 
effectiveness in enhancing the robustness of unsupervised 
models against adversarial attacks. Secondly, we apply 
and validate this training method with $k$-means clustering. 
We note the potential extension of this method to other 
unsupervised learning techniques. Finally, we highlight 
the effectiveness of transferability by utilising a 
supervised model in targeting an unsupervised model, 
with both trained on different datasets.

\section{Background}\label{prelim}

In this section we introduce the machine learning 
concepts of adversarial examples and adversarial 
training. With the latter representing an effective 
defence strategy against the former. 

\subsection{Adversarial Examples}\label{problem}

Given a standard clustering task, let $x$ 
be an image and $g$ be a clustering model. 
An adversarial example to $g$ can be crafted 
through solving the following optimisation 
problem:

\begin{equation}
\min_{\delta} d(x, x + \delta) \; \text{such 
that} \; g(x) \neq g(x + \delta),
\end{equation}

\noindent where $d$ measures similarity. 
This optimisation problem searches for a minimal 
perturbation $\delta$ that can change the class 
assignment for $x$ or expected output of the 
model \cite{dalvi2004adversarial}.

Depending on $g$'s application, the adversarial 
example $x' = x + \delta$ can have devastating 
effects \cite{martins2020adversarial, lin2021adversarial, 
apruzzese2023real}. Moreover, in some instances, 
$x'$ can be somewhat model-agnostic, such that when 
generated for model $g$, it can be effective in 
fooling another model $f$, which is either different 
in architecture, training dataset or both 
\cite{demontis2019adversarial}. We exploit this exact 
phenomenon in generating adversarial examples for 
$k$-means clustering.

\subsection{Adversarial Training} 

Mitigating the effects of adversarial examples
commonly involves adversarial training. This 
defence strategy utilises adversarial examples during 
training to improve a model's performance on 
similar examples during deployment. Existing 
adversarial training methods primarily focus 
on supervised learning approaches. This defence 
strategy can be formulated as a minimax 
optimisation problem \cite{madry2017towards}:

\begin{equation}
\min_{\theta} \mathbb{E}_{(x,y)\sim \mathcal{D}} 
[\max_{\delta \in \Delta} L(x + \delta, 
y, \theta)],
\label{adv_train}
\end{equation}

\noindent where $\Delta$ is the perturbation 
set, $L(x, y, \theta)$ is the loss function 
of a neural network with parameters $\theta$
and $(x, y)$ is the input-label pair taken from 
the distribution $\mathcal{D}$.  This minimax 
problem is often solved by first crafting adversarial 
examples to solve the loss maximisation problem 
$\max_{\delta \in \Delta}L(x + \delta, y, \theta)$ 
and then optimising the model parameters $\theta$ 
using the generated adversarial examples.

\section{Methodology}\label{method}

In this section, we detail our use of transferability.
We also detail the specifics of our proposed adversarial 
training algorithm and its application to $k$-means 
clustering.



\subsection{Exploiting Transferability}

Generally, Eq. (\ref{adv_train}) cannot be 
directly applied to unsupervised algorithms 
due to its dependence on label $y$. However, this 
equation can be approximated by either replacing 
$y$ with $g(x)$ or $f(x)$ such that $f(x) \approx 
g(x)$ and $\theta = \theta_f$, with $f$ representing 
the surrogate or substitute model, $g$ the target 
model (e.g., $k$-means) and $g(x)$ a cluster 
identifier. The input pair $(x, y)$ then becomes 
$(x, g(x))$ in the absence of a substitute model, 
or conversely, $(x, f(x))$.

Having $f(x) \approx g(x)$ can be achieved 
by constructing $f$ such that the outputs of 
$f$ are similar in dimension to the outputs 
of $g$. Thereafter $f$ can be trained on the 
outputs of $g$ for a given input $x$. Typically 
in traditional adversarial training, $f$ is a 
neural network designed to approximate the 
behaviour of the target model $g$.


In cases where labels are readily available for 
the datasets, it is hypothetically possible to 
use a pre-trained supervised model as a surrogate 
model $f$ to approximate the loss function $L(x 
+ \delta, y, \theta_f)$ for an unsupervised target 
model $g$. In such a situation, the concept of 
transferability may be leveraged to generate 
adversarial examples $x + \delta$ for model $f$ 
and utilised to enhance the robustness of model 
$g$. This situation exploits the fact that 
adversarial examples crafted for one model 
can be effective against other models 
\cite{demontis2019adversarial}, and that most 
machine learning models rely on inductive bias 
\cite{mitchell1980need, gordon1995evaluation}. 
In this paper, we try to realise and demonstrate 
this hypothesis.

\subsection{Improving Robustness}\label{training} 

For the purposes of realising and demonstrating 
the hypothesis above, we use ground-truth label 
$y$ in solving Eq. (\ref{adv_train}). Despite the 
presence of $y$, target model $g$ is trained in 
an unsupervised manner and the ground-truth label 
is not required for the proposed adversarial 
training method to function.

To ensure that the target model $g$, (i.e., 
$k$-means), maintains competitive accuracies on 
both clean and adversarial data with the proposed 
training method, we manipulate the proportion $\eta$ 
of clean and adversarial examples in the training 
dataset $\mathcal{D}'$. Furthermore, we increment 
training attack strength $\epsilon$ values at each 
step $s$ with $\epsilon = s/\beta$, where $\beta$ 
is the maximum number of steps or alternatively 
referred to as the adversarial step-count, and train 
$k$-means on both clean and adversarial examples 
in $\mathcal{D}'$. In each training step $s$ we 
anchor clusters by initialising each step with the 
centroids from the previous step $\mathtt{centroids
}_{s-1}$. After training, the final centroids $
\mathtt{centroids}_\beta$ can be utilised appropriately 
in the required application. For more details, see 
Algorithm \ref{alg2:defend} below.

\begin{algorithm}
\caption{$k$-means Adversarial Training}
\begin{algorithmic}[1]
\Input training dataset $\mathcal{D}$, number 
of clusters $k$, proportion size $\eta$, 
surrogate model $f(\cdot)$, adversarial step-count 
$\beta$
\Output $k\texttt{-}\mathtt{means.CENTROIDS}$ 
\State \textbf{initialise}: $\mathtt{centroids} = 
k\texttt{-}\mathtt{means}^{++}(\mathcal{D}, k)$
\State \textbf{initialise}: $A = \mathtt{random.uniform}
(\mathcal{D}, \vert \mathcal{D} \vert \times \eta)$
\State \textbf{initialise}: $B = \mathcal{D} - A$
\State $k\texttt{-}\mathtt{means.TRAIN}(\mathcal{D})$
\For {each $s$ in $[1, \ldots, \beta]$}
	\State $\epsilon = s/\beta$
	\State $A' = \mathtt{attack}(A, f, \epsilon) \qquad 
	\qquad \qquad \triangleright$ apply I-FGSM
	\State $\mathcal{D}' = A' \cup B$
	\State $k\texttt{-}\mathtt{means.TRAIN}(\mathcal{D}')$
\EndFor
\end{algorithmic}
\label{alg2:defend}
\end{algorithm} 

\section{Methods and Resources}
In this section, we detail the datasets and attacks 
utilised in our experiments. We also provide the 
necessary implementation details for replication. 

\subsection{Datasets}\label{dataset}

To evaluate our method, we utilise the MNIST 
\cite{lecun1998mnist} and Fashion-MNIST 
\cite{xiao2017fashion} datasets. MNIST and 
Fashion-MNIST each contain a total of 70,000 
samples, with 60,000 training and 10,000 testing. 
In both sets, each sample is a $28 \times 28$ 
monochrome 0 to 255 normalised image. MNIST 
contains handwritten digits, while Fashion-MNIST 
consists of various fashion products, (see Fig. 
\ref{collages}). These datasets are primarily chosen 
due to their use in benchmarking, but also due to 
their simplicity, accessibility, popularity and 
resource-efficiency. 

\subsection{Adversarial Attacks}\label{attacks}

For attacks, we utilise the iterative Fast Gradient 
Sign Method (I-FGSM). This is one of many $\ell_\infty$ 
attacks that can be substituted into our training 
method. I-FGSM is an iterative version of the Fast 
Gradient Sign Method (FGSM), which operates by 
adjusting the input data in an attempt to maximise loss 
at each adversarial step. For more details on the 
attack algorithm, we refer the interested reader to 
\cite{madry2017towards}. 

\subsection{Implementation Details}\label{implement}

For software, we use Python 3.9 and we implement 
our adversarial attacks with the Adversarial-Robustness-Toolbox 
\cite{art2018}. The attack parameters for I-FGSM 
are set to default, except for the epsilon step-size 
$\alpha$, which is set to $\epsilon/4$. We implement 
$k$-means with the scikit-learn Python library 
\cite{pedregosa2011scikit}. To determine the number 
of clusters $k$ for $k$-means, we use the elbow 
method heuristic \cite{yuan2019research}. In all 
our experiments we use $k = 856$ and translate the 
clusters into classes with majority voting to 
calculate clustering accuracy relative to the 
ground-truth labels. For the surrogate model $f$, 
we use ResNet-18 \cite{he2016deep} trained on the 
MNIST dataset. For adversarial training parameters, 
we use proportion size $\eta = 1/2$ and adversarial 
training step-count $\beta = 40$, unless stated 
otherwise. To evaluate the performance of the 
clustering algorithm under attack, we set $\epsilon 
= 1$ for attacks on testing data and report clustering 
accuracy. We repeat the experiments 30 times and 
report the average results. For hardware, we conduct 
our experiments on an Amazon Web Services cloud 
computer containing an NVIDIA Tesla T4 16GB 585MHz 
GPU and an Intel Xeon Platinum 8259CL 16 Cores 2.50GHz 
CPU.

\section{Results and Discussion}\label{results}

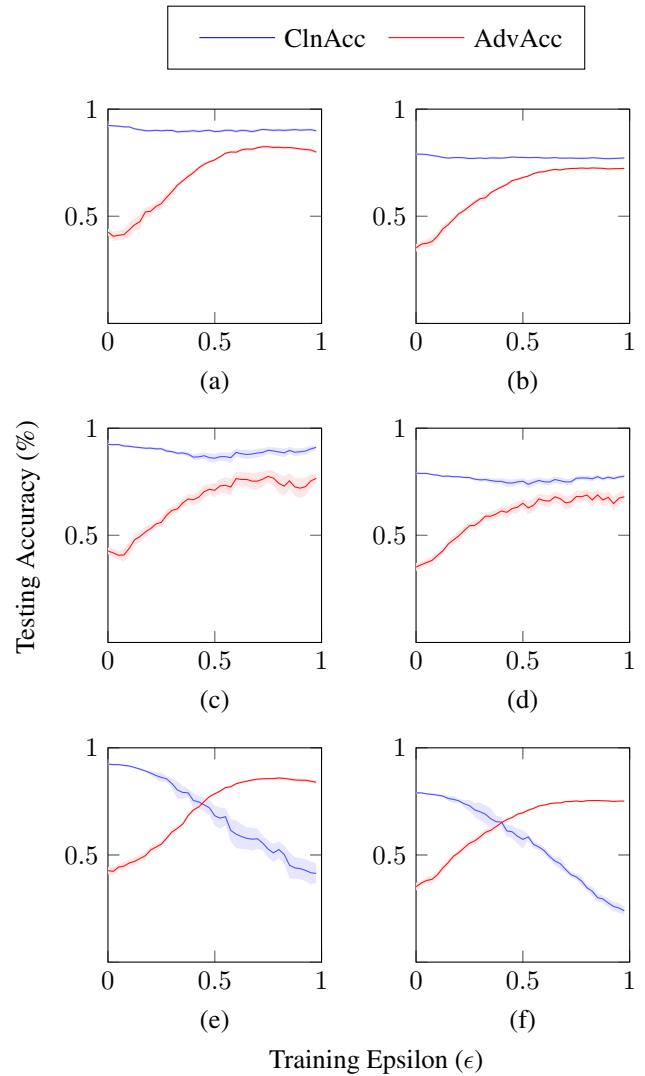
\begin{figure}[!t]
\centering
\subfloat{}\hspace{3.5em}
\subfloat{
\begin{tikzpicture}
  \node (A){};
  \node [right=of A](cln){ClnAcc};
  \node [right=of A](B){};
  \node [right=of B](C){};
  \node [right=of C](D){};
  \node [right=of C](adv){AdvAcc};
  \draw[blue] (A) -- (B);
  \draw[red] (C) -- (D);
  
  \draw[] ([shift={(-5pt,5pt)}]current bounding box.north west) rectangle ([shift={(5pt,-5pt)}]current bounding box.south east);
\end{tikzpicture}}\\\vspace*{-0.2em}
\subfloat{
\begin{tikzpicture}[rotate=90,transform shape]
  \node (A){$\qquad$ \textcolor{white}{Testing Accuracy (\%)}};
\end{tikzpicture}}
\subfloat{
\begin{tikzpicture}
\begin{axis}[%
		name=plot1,
   		table/col sep=comma,  	
        width=4.4cm,
        height=4.4cm,
        try min ticks=3,
        xticklabels={},yticklabels={},
        extra x ticks={0,0.5,1.0},extra y ticks={0.5,1.0},
        ymin=0,ymax=1.0,
        xmin=0,xmax=1.0,
        ]
		\addplot[mark=none,blue!70!white] table [x=Epsilon, 
        y=CleanTestAvg] {data/standard/mnist_fgm_fgm.csv};
		\addplot [name path=au,draw=none] table[x=Epsilon,y 
		expr=\thisrow{CleanTestAvg}+\thisrow{ClnError}] 
		{data/standard/mnist_fgm_fgm.csv};
		\addplot [name path=al,draw=none] table[x=Epsilon,y
		expr=\thisrow{CleanTestAvg}-\thisrow{ClnError}] 
		{data/standard/mnist_fgm_fgm.csv};
		\addplot [fill=blue!10] fill between[of=au and al]; 
        
  		\addplot[mark=none,red!95!white] table [x=Epsilon, 
  		y=AdvTestAvg] {data/standard/mnist_fgm_fgm.csv};
  		\addplot [name path=bu,draw=none] table[x=Epsilon,y 
		expr=\thisrow{AdvTestAvg}+\thisrow{AdvError}] 
		{data/standard/mnist_fgm_fgm.csv};\
		\addplot [name path=bl,draw=none] table[x=Epsilon,y 
		expr=\thisrow{AdvTestAvg}-\thisrow{AdvError}] 
		{data/standard/mnist_fgm_fgm.csv};
		\addplot [fill=red!10] fill between[of=bu and bl];
\end{axis}
\node[below = 0.5cm of plot1] {(a)};
\end{tikzpicture}}
\hspace*{0.5em}
\subfloat{
\begin{tikzpicture}
\begin{axis}[%
		name=plot2,
   		table/col sep=comma,  	
        width=4.4cm,
        height=4.4cm,
        try min ticks=3,
        xticklabels={},yticklabels={},
        extra x ticks={0,0.5,1.0},extra y ticks={0.5,1.0},
        ymin=0,ymax=1.0,
        xmin=0,xmax=1.0,
        ]
		\addplot[mark=none,blue!70!white] table [x=Epsilon, 
        y=CleanTestAvg] {data/standard/fashion_fgm_fgm.csv};
		\addplot [name path=cu,draw=none] table[x=Epsilon,y 
		expr=\thisrow{CleanTestAvg}+\thisrow{ClnError}] 
		{data/standard/fashion_fgm_fgm.csv};
		\addplot [name path=cl,draw=none] table[x=Epsilon,y
		expr=\thisrow{CleanTestAvg}-\thisrow{ClnError}] 
		{data/standard/fashion_fgm_fgm.csv};
		\addplot [fill=blue!10] fill between[of=cu and cl]; 
        
  		\addplot[mark=none,red!95!white] table [x=Epsilon, 
  		y=AdvTestAvg] {data/standard/fashion_fgm_fgm.csv};
  		\addplot [name path=au,draw=none] table[x=Epsilon,y 
		expr=\thisrow{AdvTestAvg}+\thisrow{AdvError}] 
		{data/standard/fashion_fgm_fgm.csv};\
		\addplot [name path=al,draw=none] table[x=Epsilon,y 
		expr=\thisrow{AdvTestAvg}-\thisrow{AdvError}] 
		{data/standard/fashion_fgm_fgm.csv};
		\addplot [fill=red!10] fill between[of=au and al];
\end{axis}
\node[below = 0.5cm of plot2] {(b)};
\end{tikzpicture}}\\\vspace*{-0.7em}

\subfloat{
\begin{tikzpicture}[rotate=90,transform shape]
  \node (A){$\; \; \; \; \; \; \;$ Testing Accuracy (\%)};
\end{tikzpicture}}
\subfloat{
\begin{tikzpicture}
\begin{axis}[%
		name=plot3,
   		table/col sep=comma,  	
        width=4.4cm,
        height=4.4cm,
        try min ticks=3,
        xticklabels={},yticklabels={},
        extra x ticks={0,0.5,1.0},extra y ticks={0.5,1.0},
        ymin=0,ymax=1.0,
        xmin=0,xmax=1.0,
        ]
		\addplot[mark=none,blue!70!white] table [x=Epsilon, 
        y=CleanTestAvg] {data/ablation/mnist_fgm_fgm_kmeans_nocentroids_ablation_pretrain_30000_samples.csv};
		\addplot [name path=cu,draw=none] table[x=Epsilon,y 
		expr=\thisrow{CleanTestAvg}+\thisrow{ClnError}] 
		{data/ablation/mnist_fgm_fgm_kmeans_nocentroids_ablation_pretrain_30000_samples.csv};
		\addplot [name path=cl,draw=none] table[x=Epsilon,y
		expr=\thisrow{CleanTestAvg}-\thisrow{ClnError}] 
		{data/ablation/mnist_fgm_fgm_kmeans_nocentroids_ablation_pretrain_30000_samples.csv};
		\addplot [fill=blue!10] fill between[of=cu and cl]; 
        
  		\addplot[mark=none,red!95!white] table [x=Epsilon, 
  		y=AdvTestAvg] {data/ablation/mnist_fgm_fgm_kmeans_nocentroids_ablation_pretrain_30000_samples.csv};
  		\coordinate (top) at (rel axis cs:0,1);
  		\addplot [name path=au,draw=none] table[x=Epsilon,y 
		expr=\thisrow{AdvTestAvg}+\thisrow{AdvError}] 
		{data/ablation/mnist_fgm_fgm_kmeans_nocentroids_ablation_pretrain_30000_samples.csv};\
		\addplot [name path=al,draw=none] table[x=Epsilon,y 
		expr=\thisrow{AdvTestAvg}-\thisrow{AdvError}] 
		{data/ablation/mnist_fgm_fgm_kmeans_nocentroids_ablation_pretrain_30000_samples.csv};
		\addplot [fill=red!10] fill between[of=au and al];
\end{axis}
\node[below = 0.5cm of plot3] {(c)};
\end{tikzpicture}}
\hspace*{0.5em}
\subfloat{
\begin{tikzpicture}
\begin{axis}[%
		name=plot4,
   		table/col sep=comma,  	
        width=4.4cm,
        height=4.4cm,
        try min ticks=3,
        xticklabels={},yticklabels={},
        extra x ticks={0,0.5,1.0},extra y ticks={0.5,1.0},
        ymin=0,ymax=1.0,
        xmin=0,xmax=1.0,
        ]
		\addplot[mark=none,blue!70!white] table [x=Epsilon, 
        y=CleanTestAvg] {data/ablation/fashion_fgm_fgm_kmeans_nocentroids_ablation_pretrain_30000_samples.csv};
		\addplot [name path=cu,draw=none] table[x=Epsilon,y 
		expr=\thisrow{CleanTestAvg}+\thisrow{ClnError}] 
		{data/ablation/fashion_fgm_fgm_kmeans_nocentroids_ablation_pretrain_30000_samples.csv};
		\addplot [name path=cl,draw=none] table[x=Epsilon,y
		expr=\thisrow{CleanTestAvg}-\thisrow{ClnError}] 
		{data/ablation/fashion_fgm_fgm_kmeans_nocentroids_ablation_pretrain_30000_samples.csv};
		\addplot [fill=blue!10] fill between[of=cu and cl]; 
        
  		\addplot[mark=none,red!95!white] table [x=Epsilon, 
  		y=AdvTestAvg] {data/ablation/fashion_fgm_fgm_kmeans_nocentroids_ablation_pretrain_30000_samples.csv};
  		\coordinate (top) at (rel axis cs:0,1);
  		\addplot [name path=au,draw=none] table[x=Epsilon,y 
		expr=\thisrow{AdvTestAvg}+\thisrow{AdvError}] 
		{data/ablation/fashion_fgm_fgm_kmeans_nocentroids_ablation_pretrain_30000_samples.csv};\
		\addplot [name path=al,draw=none] table[x=Epsilon,y 
		expr=\thisrow{AdvTestAvg}-\thisrow{AdvError}] 
		{data/ablation/fashion_fgm_fgm_kmeans_nocentroids_ablation_pretrain_30000_samples.csv};
		\addplot [fill=red!10] fill between[of=au and al];
\end{axis}
\node[below = 0.5cm of plot4] {(d)};
\end{tikzpicture}}\\\vspace*{-0.7em}

\subfloat{
\begin{tikzpicture}[rotate=90,transform shape]
  \node (A){$\qquad$ \textcolor{white}{Testing Accuracy (\%)}};
\end{tikzpicture}}
\subfloat{
\begin{tikzpicture}
\begin{axis}[%
		name=plot5,
   		table/col sep=comma,  	
        width=4.4cm,
        height=4.4cm,
        try min ticks=3,
        xticklabels={},yticklabels={},
        extra x ticks={0,0.5,1.0},extra y ticks={0.5,1.0},
        ymin=0,ymax=1.0,
        xmin=0,xmax=1.0,
        ]
		\addplot[mark=none,blue!70!white] table [x=Epsilon, 
        y=CleanTestAvg] {data/ablation/mnist_fgm_fgm_kmeans_60000_samples.csv};
		\addplot [name path=cu,draw=none] table[x=Epsilon,y 
		expr=\thisrow{CleanTestAvg}+\thisrow{ClnError}] 
		{data/ablation/mnist_fgm_fgm_kmeans_60000_samples.csv};
		\addplot [name path=cl,draw=none] table[x=Epsilon,y
		expr=\thisrow{CleanTestAvg}-\thisrow{ClnError}] 
		{data/ablation/mnist_fgm_fgm_kmeans_60000_samples.csv};
		\addplot [fill=blue!10] fill between[of=cu and cl]; 
        
  		\addplot[mark=none,red!95!white] table [x=Epsilon, 
  		y=AdvTestAvg] {data/ablation/mnist_fgm_fgm_kmeans_60000_samples.csv};
  		\coordinate (top) at (rel axis cs:0,1);
  		\addplot [name path=au,draw=none] table[x=Epsilon,y 
		expr=\thisrow{AdvTestAvg}+\thisrow{AdvError}] 
		{data/ablation/mnist_fgm_fgm_kmeans_60000_samples.csv};\
		\addplot [name path=al,draw=none] table[x=Epsilon,y 
		expr=\thisrow{AdvTestAvg}-\thisrow{AdvError}] 
		{data/ablation/mnist_fgm_fgm_kmeans_60000_samples.csv};
		\addplot [fill=red!10] fill between[of=au and al];
\end{axis}
\node[below = 0.5cm of plot5] {(e)};
\end{tikzpicture}}
\hspace*{0.5em}
\subfloat{
\begin{tikzpicture}
\begin{axis}[%
		name=plot6,
   		table/col sep=comma,  	
        width=4.4cm,
        height=4.4cm,
        try min ticks=3,
        xticklabels={},yticklabels={},
        extra x ticks={0,0.5,1.0},extra y ticks={0.5,1.0},
        ymin=0,ymax=1.0,
        xmin=0,xmax=1.0,
        ]
		\addplot[mark=none,blue!70!white] table [x=Epsilon, 
        y=CleanTestAvg] {data/ablation/fashion_fgm_fgm_kmeans_60000_samples.csv};
		\addplot [name path=cu,draw=none] table[x=Epsilon,y 
		expr=\thisrow{CleanTestAvg}+\thisrow{ClnError}] 
		{data/ablation/fashion_fgm_fgm_kmeans_60000_samples.csv};
		\addplot [name path=cl,draw=none] table[x=Epsilon,y
		expr=\thisrow{CleanTestAvg}-\thisrow{ClnError}] 
		{data/ablation/fashion_fgm_fgm_kmeans_60000_samples.csv};
		\addplot [fill=blue!10] fill between[of=cu and cl]; 
        
  		\addplot[mark=none,red!95!white] table [x=Epsilon, 
  		y=AdvTestAvg] {data/ablation/fashion_fgm_fgm_kmeans_60000_samples.csv};
  		\coordinate (top) at (rel axis cs:0,1);
  		\addplot [name path=au,draw=none] table[x=Epsilon,y 
		expr=\thisrow{AdvTestAvg}+\thisrow{AdvError}] 
		{data/ablation/fashion_fgm_fgm_kmeans_60000_samples.csv};\
		\addplot [name path=al,draw=none] table[x=Epsilon,y 
		expr=\thisrow{AdvTestAvg}-\thisrow{AdvError}] 
		{data/ablation/fashion_fgm_fgm_kmeans_60000_samples.csv};
		\addplot [fill=red!10] fill between[of=au and al];
\end{axis}
\node[below = 0.5cm of plot6] {(f)};
\end{tikzpicture}}\\\vspace*{-1em}
\subfloat{}\hspace{3.5em}
\subfloat{
\begin{tikzpicture}
  \node (A){Training Epsilon ($\epsilon$)};
\end{tikzpicture}}\\
\caption{Clean (ClnAcc) and adversarial (AdvAcc) 
clustering accuracies, on MNIST and Fashion-MNIST. 
For all the provided plots we have I-FGSM as the 
attack and adversarial training algorithm, the 
attack strength ($\epsilon$) used in training 
along the $x$-axis and the clustering accuracies 
($\%$) along the $y$-axis. In each plot, the 
solid lines represent the average results from 
30 experiments, while shaded areas illustrate 
the error bars for a confidence level of $99\%$. 
In the first column we have results for MNIST 
and Fashion-MNIST in the second. In (a) and (b) 
we have the full implementation of the proposed 
adversarial training algorithm. In (c) to (f) 
we have parameter sensitivity results. In (c) 
and (d) we have $k$-means trained in a similar 
manner as to that in (a) and (b), however without 
the initialisation of centroids from previous 
steps, i.e., no continuous learning. In (e) and 
(f) we have fully perturbed training sets as 
opposed to half of the training sets, i.e., $\eta=1$}
\label{plots}
\end{figure}

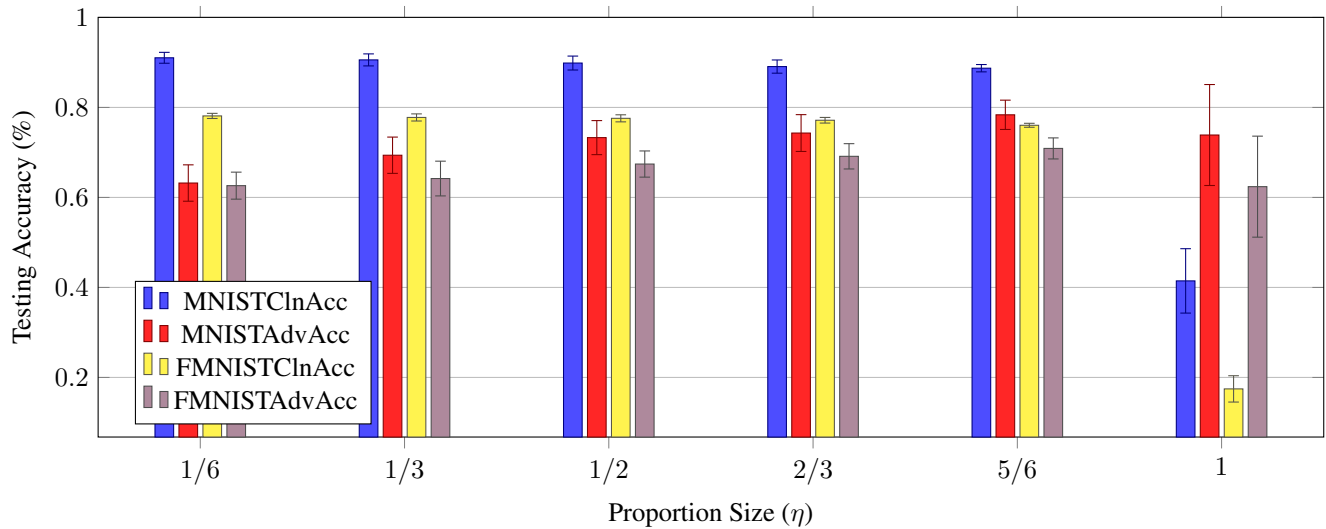
\begin{figure*}[!t]
\begin{tikzpicture}
\begin{axis}[
	table/col sep=comma,	
	width=1\textwidth,
	height=.4\textwidth,
    ybar,
    bar width=7pt,
    xlabel={Proportion Size ($\eta$)},
    ylabel={Testing Accuracy (\%)},
    xtick=data,
    xticklabels={$1/6$, $1/3$, $1/2$, $2/3$, $5/6$, $1$},
    ymajorgrids,
    legend pos=south west,
    error bars/y dir=both, 
	error bars/y explicit  
    ]
    \addplot[blue!50!black,fill=blue!70!white] table [x expr=\coordindex, y=CleanTestAvg,
    y error=ClnError]{data/ablation/mnist_fgm_fgm_kmeans_ablation_datasize_collated.csv};
    \addlegendentry{MNISTClnAcc}
    \addplot[red!50!black,fill=red!85!white] table [x expr=\coordindex, y=AdvTestAvg, 
    y error=AdvError]{data/ablation/mnist_fgm_fgm_kmeans_ablation_datasize_collated.csv};
    \addlegendentry{MNISTAdvAcc}
    
    \addplot[yellow!20!black,fill=yellow!80!white] table [x expr=\coordindex, y=CleanTestAvg,
    y error=ClnError]{data/ablation/fashion_fgm_fgm_kmeans_ablation_datasize_collated.csv};
    \addlegendentry{FMNISTClnAcc}
    \addplot[gray!60!black,fill=gray!70!magenta] table [x expr=\coordindex, y=AdvTestAvg, 
    y error=AdvError]{data/ablation/fashion_fgm_fgm_kmeans_ablation_datasize_collated.csv};
    \addlegendentry{FMNISTAdvAcc}
\end{axis}
\end{tikzpicture}
\caption{Parameter sensitivity results on various 
adversarial to clean proportions, without incremental 
training, i.e., different values for $\eta$ and when 
adversarial step-count $\beta=1$. Along the $x$-axis 
we have proportion size $\eta$, used in controlling 
the ration between clean and adversarial data. Along 
the $y$-axis we have the clustering accuracies ($\%$). 
Each shaded bar illustrates average testing accuracies 
on MNIST and Fashion-MNIST after 30 experiments. Error 
bars are for a confidence level of $99\%$. I-FGSM is 
the attacking and defending algorithm. For all 
proportions, both training and testing attack strengths 
use $\epsilon = 1$.}
\label{sizes}
\end{figure*}

Here we evaluate the performance of $k$-means 
clustering under adversarial attacks and 
the effectiveness of our adversarial training
algorithm. As demonstrated in Fig. \ref{plots}, 
several trends and patterns emerge. Each shedding 
light on the impact of epsilon values on 
accuracy, the behaviour of accuracy with or 
without continuous learning, and the relationship 
between data distribution, robustness and clean 
performance.

One notable observation from Fig. \ref{plots}
is the effectiveness of transferability. 
Transferability, in machine learning, refers to 
the ability of adversarial attacks generated for 
one model or dataset to successfully impact the 
performance of a different model or dataset. This 
concept has been studied extensively in previous 
work for evasion attacks, e.g., \cite{dong2018boosting, 
liu2016delving, papernot2016transferability, 
papernot2017practical, szegedy2013intriguing, wang2023role}. 
Notably, Biggio et al. \cite{biggio2013evasion} 
were the first to consider evasion attacks against 
surrogate models in a limited-knowledge scenario, 
while Goodfellow et al. \cite{goodfellow2014explaining}, 
Tramer et al. \cite{tramer2017space}, and Moosavi 
et al. \cite{moosavi2017universal} were some of 
the first to make the observation that different 
models might learn intersecting decision boundaries 
in both benign and adversarial dimensions. In 
practical  scenarios, adversarial attacks often 
exploit surrogate models due to limited access 
to the target model's architecture or loss function 
\cite{papernot2017practical}. To our knowledge 
however, the work presented here is the first 
example of a supervised model being utilised as 
a surrogate model in targeting an unsupervised 
model and a surprising example of transferability 
working across two different datasets.

Namely, in all our experiments, the surrogate 
model $f$ was a ResNet-18 model trained on the 
MNIST handwritten digits dataset. The model 
did not receive any further training or tuning. 
The generated attacks on Fashion-MNIST for model 
$g$ ($k$-means) were generated with $f$, without 
any further training on Fashion-MNIST. That is, 
transferability was exploited explicitly in crafting 
adversarial examples using a supervised ResNet-18 
model, which was trained on the MNIST dataset, to 
attack an unsupervised $k$-means clustering model 
on the Fashion-MNIST dataset. Considering this, 
we observe some very surprising results in Fig. 
\ref{plots}(a-f). We observe that each row illustrates 
strikingly similar behaviour, where plots in the 
first column illustrate performance on MNIST and 
the second on Fashion-MNIST. We note that any 
observable differences in performance between MNIST 
and Fashion-MNIST (e.g., Fig. \ref{plots}(c) and 
\ref{plots}(d)) may be a result of MNIST being 
a simpler dataset than Fashion-MNIST. That 
is, any difference in performance may not be 
strictly due to the quality or efficacy of the 
adversarial examples generated with surrogate 
model $f$.

In Fig. \ref{plots}(a) and \ref{plots}(b) we 
also observe how epsilon values affect accuracy 
metrics. As $\epsilon$ increases, we notice a 
consistent decline in accuracy on clean test data, 
i.e., from approximately 92\% to 90\% for MNIST (Fig. 
\ref{plots}(a)), and 79\% to 78\% for Fashion-MNIST 
(Fig. \ref{plots}(b)). Conversely, accuracy 
on adversarial test data shows a gradual 
increase with each increase in epsilon, i.e., 
from 43\% to 80\% for MNIST, and 38\% to 
74\% for Fashion-MNIST. This behaviour demonstrates 
a well-established trade-off between robustness 
against adversarial attacks and performance 
on clean data. As models are made more robust, 
their performance on clean data tends to degrade. 
However in our case, we only witness slight 
degradation.

Switching focus to our parameter sensitivity studies, 
in Fig. \ref{plots}(c-f), we observe the effects 
of two key parameters of our adversarial training 
algorithm. Comparing Fig. \ref{plots}(a) and 
\ref{plots}(c) we observe the importance of initialising 
the centroids of each step $\mathtt{centroids}_s$ 
with centroids from the previous step $\mathtt{
centroids}_{s-1}$, thereby anchoring learnt clusters 
and establishing continuous learning. The same 
observation can be made for Fashion-MNIST when 
comparing Fig. \ref{plots}(b) and \ref{plots}(d), 
where the continuous learning strategy clearly 
stabilises both clean and adversarial testing 
accuracy. In Fig. \ref{plots}(e) and \ref{plots}(f) 
we observe the importance of having an even 
proportion of clean and adversarial examples. 
In Fig. \ref{plots}(e) and \ref{plots}(f) we have 
$\eta = 1$ as opposed to $\eta = 1/2$ as in Fig. 
\ref{plots}(a) and \ref{plots}(b).

The importance of controlling clean and adversarial 
proportions with $\eta$ is further highlighted 
in Fig. \ref{sizes}. Here we observe that a 
proportion of $\eta= 1/2$ or $\eta = 2/3$ 
adversarial examples results in competitive 
performance for both datasets, with the former 
having slightly better clean test accuracy (i.e., 
89.9\% vs. 89.1\% for MNIST and 77.6\% vs. 77.2\% 
for Fashion-MNIST) and the latter having slightly 
better adversarial testing accuracy (i.e., 73.3\% 
vs. 74.3\% for MNIST and 67.4\% vs. 69.1\% for 
Fashion-MNIST). A proportion of $\eta = 5/6$ 
adversarial examples results in the best adversarial 
testing accuracy (78.4\% for MNIST and 70.9\% for 
Fashion-MNIST), however at the expense of clean 
testing accuracy (i.e., 88.7\% for MNIST and 
76.0\% for Fashion-MNIST). For our purposes, the 
choice between $\eta= 1/2$, $\eta= 2/3$ and $\eta= 
5/6$ considered both time efficiency and performance.
$\eta= 1/2$ resulted in the least amount of time 
and training data required for acceptable performance, 
especially when statistically significant replications 
of each experiment had to be conducted.

Continuing with further observations from Fig. 
\ref{sizes}, we see that when $\eta = 1$, we 
have the worst clean (17.5\%) and adversarial 
testing (62.4\%) performance for Fashion-MNIST and 
acceptable performance for MNIST (i.e., 41.5\% and 
73.9\%, respectively). For both datasets, we observe 
the greatest amount of variance when $\eta = 1$. 
We also coincidentally observe the importance of 
the step-count parameter $\beta$, i.e., incremental 
training. Namely when $\eta = 1$ and incremental 
training is absent, the adversarial testing 
accuracy is relatively acceptable for MNIST 
(i.e., 73.9\%) and considerably low for Fashion-MNIST 
(62.4\%). Conversely, when the entire training 
set is perturbed with an adversarial step-count 
of $\beta = 40$, i.e., when incremental training 
is available, the $k$-means algorithm attains an 
adversarial testing accuracy of 84\% on MNIST and 
75\% on Fashion-MNIST, as shown in Fig. \ref{plots}(e) 
and \ref{plots}(f) respectively. 

These results generally highlight the importance 
of continuous learning via centroid initialisation 
and emphasise the sensitivity of unsupervised models 
to sample distributions. For sensitivity in particular, 
unsupervised algorithms, unlike their supervised 
counterparts, operate without reliance on labelled 
data or environmental signals \cite{darling1970non, 
ghahramani2003unsupervised}. Hence sample distributions 
determine equitable or non-equitable exposure, either 
enhancing or degrading an algorithm's performance in 
recognising and characterising underlying patterns 
\cite{qian2013robust}.

Before concluding, we emphasise that the work 
presented here is a special case of our proposed 
adversarial training method. In a strict unsupervised 
scenario, the input pair $(x, y)$ in Eq. \ref{adv_train} 
must be approximated. In the case where a 
label-independent attack exists (e.g., evolutionary 
attack \cite{su2019one} with modification) and 
target model $g$ has an accessible loss function 
$L$, then the input pair can be approximated with 
$(x, g(x))$, where $g(x)$ is a cluster identifier. 
Otherwise in the case where $g$'s loss function is 
not accessible or expressible, then a surrogate 
model $f$ must be constructed to approximate $g$, 
such that $f(x) \approx g(x)$, and the input pair 
$(x, y)$ can be replaced with $(x, f(x))$ and model 
parameters $\theta$ with $\theta_f$. Additionally, 
as presented in this work, if ground-truth label 
$y$ is available but an appropriate attack or loss 
function $L$ for $g$ is not readily available, then 
the model parameters $\theta$ are replaced with 
$\theta_f$. If $y$ and $g(x)$ have different dimensions, 
then a post-processing step such as majority voting 
or the Hungarian method can be utilised in resolving 
this assignment problem \cite{kuhn1955hungarian}.

Due to the effectiveness of transferability, 
we consider it immaterial whether the substitute 
model $f$ is a pre-trained model or not. Our focus 
on pre-trained models is a result of their ubiquity 
and their permissibility as vectors in generating 
and transferring adversarial attacks. The choice 
of datasets follows a similar vein, i.e., the MNIST 
and Fashion-MNIST datasets are accessible, popular 
and are often used for benchmarking. The two datasets 
also contain features that are representative of 
broader image datasets, enabling results generated 
from these datasets to be potentially applicable to 
practical scenarios involving image analysis and 
clustering (e.g., image segmentation and anomaly 
detection). Additionally, since MNIST and Fashion-MNIST 
have relatively limited feature spaces and homogeneous 
backgrounds compared to more complex image datasets, 
they have allowed us to study a scenario in which 
adversaries have limited ability in camouflaging 
manipulations. But more importantly, they have allowed 
us to establish one of the first baselines for 
unsupervised adversarial training.

Finally, the findings discussed here extend beyond 
image analysis and clustering, and have significant 
implications for the security and reliability 
of unsupervised learning and clustering applications 
in general. For instance, applications such as 
customer segmentation in targeted marketing for 
e-commerce platforms or disease sub-typing based on 
patient data in healthcare applications may face 
accuracy and reliability challenges amid adversarial 
modifications. Additional examples also include 
manufacturing and autonomous vehicles, where adversarial 
examples could lead to safety and quality control 
concerns if adversarial manipulations mislead the 
clustering process. It is due to these examples, 
and others, that we stress the importance of 
developing robust algorithms. 

\section{Conclusion}\label{conclusion}

Our evaluation of the performance of $k$-means 
clustering under adversarial attacks and the 
effectiveness of our adversarial training algorithm 
sheds light on several important factors. We have 
observed the impact of adversarial step-count $\beta$ 
on accuracy, the trade-off between robustness against 
adversarial attacks and performance on clean data, 
and the significance of key parameters in our 
adversarial training algorithm.

One notable finding is the effectiveness of 
transferability, which highlights the potential 
of using supervised models as surrogates for 
target unsupervised models. This finding emphasises 
the importance of considering diverse scenarios 
in model training and evaluation, since real-world 
applications often involve models facing data 
distributions not encountered during training. 

This paper also highlights the sensitivity of 
unsupervised models to sample distributions, 
emphasising the need for careful control of clean 
and adversarial example proportions. Our results 
underscore the importance of continuous learning 
and proper initialisation of centroids, which can 
significantly enhance both clean and adversarial 
testing accuracy. For practical implications, 
we consider that the insights generated in this 
study may find utility in deployed machine learning 
models, especially those utilising unsupervised 
learning or clustering algorithms.  

Overall, our study emphasises the vulnerability 
of unsupervised learning and clustering algorithms 
to adversarial attacks and provides insights into 
potential defence mechanisms. Future research 
could explore dynamic defence mechanisms that 
adapt to the specific characteristics of the 
surrogate model and the target model. Future 
studies may also investigate the incorporation 
of domain knowledge to enhance model robustness, 
i.e., understanding the inherent characteristics 
of different datasets could inform the development 
of more resilient models. Finally, extending our
method to diverse real-world applications, 
such as medical imaging or cyber-security, could 
provide a more comprehensive understanding of the 
practical implications and generalisability of 
our findings. We hope these avenues, and others, 
will contribute to the ongoing efforts to fortify 
machine learning models.

\bibliographystyle{IEEEtran}
\bibliography{paperbib}

\begin{thebibliography}{10}
\providecommand{\url}[1]{#1}
\csname url@samestyle\endcsname
\providecommand{\newblock}{\relax}
\providecommand{\bibinfo}[2]{#2}
\providecommand{\BIBentrySTDinterwordspacing}{\spaceskip=0pt\relax}
\providecommand{\BIBentryALTinterwordstretchfactor}{4}
\providecommand{\BIBentryALTinterwordspacing}{\spaceskip=\fontdimen2\font plus
\BIBentryALTinterwordstretchfactor\fontdimen3\font minus
  \fontdimen4\font\relax}
\providecommand{\BIBforeignlanguage}[2]{{%
\expandafter\ifx\csname l@#1\endcsname\relax
\typeout{** WARNING: IEEEtran.bst: No hyphenation pattern has been}%
\typeout{** loaded for the language `#1'. Using the pattern for}%
\typeout{** the default language instead.}%
\else
\language=\csname l@#1\endcsname
\fi
#2}}
\providecommand{\BIBdecl}{\relax}
\BIBdecl

\bibitem{xu2020adversarial}
H.~Xu, Y.~Ma, H.-C. Liu, D.~Deb, H.~Liu, J.-L. Tang, and A.~K. Jain,
  ``Adversarial attacks and defenses in images, graphs and text: A review,''
  \emph{International Journal of Automation and Computing}, vol.~17, pp.
  151--178, 2020.

\bibitem{chakraborty2021survey}
A.~Chakraborty, M.~Alam, V.~Dey, A.~Chattopadhyay, and D.~Mukhopadhyay, ``A
  survey on adversarial attacks and defences,'' \emph{CAAI Transactions on
  Intelligence Technology}, vol.~6, no.~1, pp. 25--45, 2021.

\bibitem{martins2020adversarial}
N.~Martins, J.~M. Cruz, T.~Cruz, and P.~H. Abreu, ``Adversarial machine
  learning applied to intrusion and malware scenarios: a systematic review,''
  \emph{IEEE Access}, vol.~8, pp. 35\,403--35\,419, 2020.

\bibitem{chhabra2020suspicion}
A.~Chhabra, A.~Roy, and P.~Mohapatra, ``Suspicion-free adversarial attacks on
  clustering algorithms,'' in \emph{Proceedings of the AAAI Conference on
  Artificial Intelligence}, vol.~34, no.~04, 2020, pp. 3625--3632.

\bibitem{dubes1980clustering}
R.~Dubes and A.~K. Jain, ``Clustering methodologies in exploratory data
  analysis,'' \emph{Advances in computers}, vol.~19, pp. 113--228, 1980.

\bibitem{khan2014dbscan}
K.~Khan, S.~U. Rehman, K.~Aziz, S.~Fong, and S.~Sarasvady, ``Dbscan: Past,
  present and future,'' in \emph{The fifth international conference on the
  applications of digital information and web technologies (ICADIWT
  2014)}.\hskip 1em plus 0.5em minus 0.4em\relax IEEE, 2014, pp. 232--238.

\bibitem{reynolds2009gaussian}
D.~A. Reynolds \emph{et~al.}, ``Gaussian mixture models.'' \emph{Encyclopedia
  of biometrics}, vol. 741, no. 659-663, 2009.

\bibitem{hartigan1979algorithm}
J.~A. Hartigan and M.~A. Wong, ``Algorithm as 136: A k-means clustering
  algorithm,'' \emph{Journal of the royal statistical society. series c
  (applied statistics)}, vol.~28, no.~1, pp. 100--108, 1979.

\bibitem{zhang1997birch}
T.~Zhang, R.~Ramakrishnan, and M.~Livny, ``Birch: A new data clustering
  algorithm and its applications,'' \emph{Data mining and knowledge discovery},
  vol.~1, pp. 141--182, 1997.

\bibitem{skillicorn2009adversarial}
D.~B. Skillicorn, ``Adversarial knowledge discovery,'' \emph{IEEE Intelligent
  Systems}, vol.~24, no.~6, p.~54, 2009.

\bibitem{biggio2013data}
B.~Biggio, I.~Pillai, S.~Rota~Bul{\`o}, D.~Ariu, M.~Pelillo, and F.~Roli, ``Is
  data clustering in adversarial settings secure?'' in \emph{Proceedings of the
  2013 ACM workshop on Artificial intelligence and security}, 2013, pp. 87--98.

\bibitem{biggio2013evasion}
B.~Biggio, I.~Corona, D.~Maiorca, B.~Nelson, N.~{\v{S}}rndi{\'c}, P.~Laskov,
  G.~Giacinto, and F.~Roli, ``Evasion attacks against machine learning at test
  time,'' in \emph{Machine Learning and Knowledge Discovery in Databases:
  European Conference, ECML PKDD 2013, Prague, Czech Republic, September 23-27,
  2013, Proceedings, Part III 13}.\hskip 1em plus 0.5em minus 0.4em\relax
  Springer, 2013, pp. 387--402.

\bibitem{crussell2015attacking}
J.~Crussell and P.~Kegelmeyer, ``Attacking dbscan for fun and profit,'' in
  \emph{Proceedings of the 2015 SIAM International Conference on Data
  Mining}.\hskip 1em plus 0.5em minus 0.4em\relax SIAM, 2015, pp. 235--243.

\bibitem{demontis2019adversarial}
A.~Demontis, M.~Melis, M.~Pintor, M.~Jagielski, B.~Biggio, A.~Oprea,
  C.~Nita-Rotaru, and F.~Roli, ``Why do adversarial attacks transfer?
  explaining transferability of evasion and poisoning attacks,'' in \emph{28th
  USENIX security symposium (USENIX security 19)}, 2019, pp. 321--338.

\bibitem{goodfellow2014explaining}
I.~J. Goodfellow, J.~Shlens, and C.~Szegedy, ``Explaining and harnessing
  adversarial examples,'' \emph{arXiv preprint arXiv:1412.6572}, 2014.

\bibitem{madry2017towards}
A.~Madry, A.~Makelov, L.~Schmidt, D.~Tsipras, and A.~Vladu, ``Towards deep
  learning models resistant to adversarial attacks,'' \emph{arXiv preprint
  arXiv:1706.06083}, 2017.

\bibitem{zhang2019theoretically}
H.~Zhang, Y.~Yu, J.~Jiao, E.~Xing, L.~El~Ghaoui, and M.~Jordan, ``Theoretically
  principled trade-off between robustness and accuracy,'' in
  \emph{International conference on machine learning}.\hskip 1em plus 0.5em
  minus 0.4em\relax PMLR, 2019, pp. 7472--7482.

\bibitem{xie2017mitigating}
C.~Xie, J.~Wang, Z.~Zhang, Z.~Ren, and A.~Yuille, ``Mitigating adversarial
  effects through randomization,'' \emph{arXiv preprint arXiv:1711.01991},
  2017.

\bibitem{guo2017countering}
C.~Guo, M.~Rana, M.~Cisse, and L.~Van Der~Maaten, ``Countering adversarial
  images using input transformations,'' \emph{arXiv preprint arXiv:1711.00117},
  2017.

\bibitem{mustafa2019image}
A.~Mustafa, S.~H. Khan, M.~Hayat, J.~Shen, and L.~Shao, ``Image
  super-resolution as a defense against adversarial attacks,'' \emph{IEEE
  Transactions on Image Processing}, vol.~29, pp. 1711--1724, 2019.

\bibitem{cohen2019certified}
J.~Cohen, E.~Rosenfeld, and Z.~Kolter, ``Certified adversarial robustness via
  randomized smoothing,'' in \emph{international conference on machine
  learning}.\hskip 1em plus 0.5em minus 0.4em\relax PMLR, 2019, pp. 1310--1320.

\bibitem{raghunathan2018certified}
A.~Raghunathan, J.~Steinhardt, and P.~Liang, ``Certified defenses against
  adversarial examples,'' \emph{arXiv preprint arXiv:1801.09344}, 2018.

\bibitem{lecun1998mnist}
``The mnist database of handwritten digits,'' \emph{http://yann. lecun.
  com/exdb/mnist/}.

\bibitem{xiao2017fashion}
H.~Xiao, K.~Rasul, and R.~Vollgraf, ``Fashion-mnist: a novel image dataset for
  benchmarking machine learning algorithms,'' \emph{arXiv preprint
  arXiv:1708.07747}, 2017.

\bibitem{dalvi2004adversarial}
N.~Dalvi, P.~Domingos, Mausam, S.~Sanghai, and D.~Verma, ``Adversarial
  classification,'' in \emph{Proceedings of the tenth ACM SIGKDD international
  conference on Knowledge discovery and data mining}, 2004, pp. 99--108.

\bibitem{lin2021adversarial}
H.-Y. Lin and B.~Biggio, ``Adversarial machine learning: Attacks from
  laboratories to the real world,'' \emph{Computer}, vol.~54, no.~5, pp.
  56--60, 2021.

\bibitem{apruzzese2023real}
G.~Apruzzese, H.~S. Anderson, S.~Dambra, D.~Freeman, F.~Pierazzi, and
  K.~Roundy, ``“real attackers don't compute gradients”: Bridging the gap
  between adversarial ml research and practice,'' in \emph{2023 IEEE Conference
  on Secure and Trustworthy Machine Learning (SaTML)}.\hskip 1em plus 0.5em
  minus 0.4em\relax IEEE, 2023, pp. 339--364.

\bibitem{mitchell1980need}
T.~M. Mitchell, ``The need for biases in learning generalizations,'' 1980.

\bibitem{gordon1995evaluation}
D.~F. Gordon and M.~Desjardins, ``Evaluation and selection of biases in machine
  learning,'' \emph{Machine learning}, vol.~20, pp. 5--22, 1995.

\bibitem{art2018}
\BIBentryALTinterwordspacing
M.-I. Nicolae, M.~Sinn, M.~N. Tran, B.~Buesser, A.~Rawat, M.~Wistuba,
  V.~Zantedeschi, N.~Baracaldo, B.~Chen, H.~Ludwig, I.~Molloy, and B.~Edwards,
  ``Adversarial robustness toolbox v1.2.0,'' \emph{CoRR}, vol. 1807.01069,
  2018. [Online]. Available: \url{https://arxiv.org/pdf/1807.01069}
\BIBentrySTDinterwordspacing

\bibitem{pedregosa2011scikit}
F.~Pedregosa, G.~Varoquaux, A.~Gramfort, V.~Michel, B.~Thirion, O.~Grisel,
  M.~Blondel, P.~Prettenhofer, R.~Weiss, V.~Dubourg \emph{et~al.},
  ``Scikit-learn: Machine learning in python,'' \emph{the Journal of machine
  Learning research}, vol.~12, pp. 2825--2830, 2011.

\bibitem{yuan2019research}
C.~Yuan and H.~Yang, ``Research on k-value selection method of k-means
  clustering algorithm,'' \emph{J}, vol.~2, no.~2, pp. 226--235, 2019.

\bibitem{he2016deep}
K.~He, X.~Zhang, S.~Ren, and J.~Sun, ``Deep residual learning for image
  recognition,'' in \emph{Proceedings of the IEEE conference on computer vision
  and pattern recognition}, 2016, pp. 770--778.

\bibitem{dong2018boosting}
Y.~Dong, F.~Liao, T.~Pang, X.~Hu, and J.~Zhu, ``Boosting adversarial examples
  with momentum,'' in \emph{Conference on Computer Vision and Pattern
  Recognition (CVPR 2018)}, 2018.

\bibitem{liu2016delving}
Y.~Liu, X.~Chen, C.~Liu, and D.~Song, ``Delving into transferable adversarial
  examples and black-box attacks,'' \emph{arXiv preprint arXiv:1611.02770},
  2016.

\bibitem{papernot2016transferability}
N.~Papernot, P.~McDaniel, and I.~Goodfellow, ``Transferability in machine
  learning: from phenomena to black-box attacks using adversarial samples,''
  \emph{arXiv preprint arXiv:1605.07277}, 2016.

\bibitem{papernot2017practical}
N.~Papernot, P.~McDaniel, I.~Goodfellow, S.~Jha, Z.~B. Celik, and A.~Swami,
  ``Practical black-box attacks against machine learning,'' in
  \emph{Proceedings of the 2017 ACM on Asia conference on computer and
  communications security}, 2017, pp. 506--519.

\bibitem{szegedy2013intriguing}
C.~Szegedy, W.~Zaremba, I.~Sutskever, J.~Bruna, D.~Erhan, I.~Goodfellow, and
  R.~Fergus, ``Intriguing properties of neural networks,'' \emph{arXiv preprint
  arXiv:1312.6199}, 2013.

\bibitem{wang2023role}
Y.~Wang and F.~Farnia, ``On the role of generalization in transferability of
  adversarial examples,'' in \emph{Uncertainty in Artificial
  Intelligence}.\hskip 1em plus 0.5em minus 0.4em\relax PMLR, 2023, pp.
  2259--2270.

\bibitem{tramer2017space}
F.~Tram{\`e}r, N.~Papernot, I.~Goodfellow, D.~Boneh, and P.~McDaniel, ``The
  space of transferable adversarial examples,'' \emph{arXiv preprint
  arXiv:1704.03453}, 2017.

\bibitem{moosavi2017universal}
S.-M. Moosavi-Dezfooli, A.~Fawzi, O.~Fawzi, and P.~Frossard, ``Universal
  adversarial perturbations,'' in \emph{Proceedings of the IEEE conference on
  computer vision and pattern recognition}, 2017, pp. 1765--1773.

\bibitem{darling1970non}
E.~M. Darling~Jr and J.~G. Raudseps, ``Non-parametric unsupervised learning
  with applications to image classification,'' \emph{Pattern Recognition},
  vol.~2, no.~4, pp. 313--335, 1970.

\bibitem{ghahramani2003unsupervised}
Z.~Ghahramani, ``Unsupervised learning,'' in \emph{Summer school on machine
  learning}.\hskip 1em plus 0.5em minus 0.4em\relax Springer, 2003, pp.
  72--112.

\bibitem{qian2013robust}
M.~Qian and C.~Zhai, ``Robust unsupervised feature selection,'' in
  \emph{Twenty-third international joint conference on artificial
  intelligence}.\hskip 1em plus 0.5em minus 0.4em\relax Citeseer, 2013.

\bibitem{su2019one}
J.~Su, D.~V. Vargas, and K.~Sakurai, ``One pixel attack for fooling deep neural
  networks,'' \emph{IEEE Transactions on Evolutionary Computation}, vol.~23,
  no.~5, pp. 828--841, 2019.

\bibitem{kuhn1955hungarian}
H.~W. Kuhn, ``The {Hungarian} method for the assignment problem,'' \emph{Naval
  research logistics quarterly}, vol.~2, no. 1-2, pp. 83--97, 1955.

\end{thebibliography}
\vskip -1.5\baselineskip plus -1fil
\begin{IEEEbiography}[{\includegraphics[width=1in,
height=1.25in,clip,keepaspectratio]{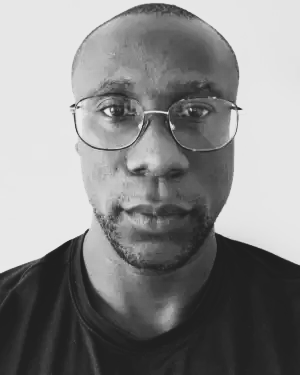}}]
{Rollin Omari}
received the B.M.S. and B.A.S (Hons) degrees from 
the University of Canberra, Canberra, Australia, in 
2014 and 2015, respectively. He is currently pursuing 
the Ph.D. degree with the College of Engineering, Computing 
and Cybernetics at the Australian National University, 
Canberra, Australia. He is also currently a researcher 
in the Defence Science Technology Group of the Australian 
Department of Defence. His current research interests 
include machine learning, bio-inspired computing and 
cyber security.
\end{IEEEbiography}  
\vskip -1.5\baselineskip plus -1fil
\begin{IEEEbiography}[{\includegraphics[width=1in,
height=1.25in,clip,keepaspectratio]{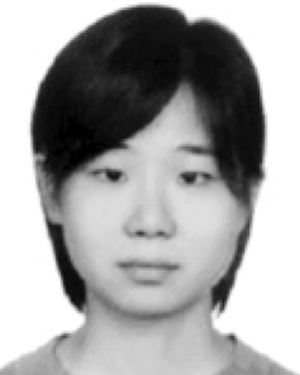}}]
{Junae Kim} Junae Kim received the B.S. degree from 
Ewha Womans University, Seoul, Korea, in 2000, the 
M.S. degree from the Pohang University of Science 
and Technology, Pohang, Korea, in 2002, the M.Phil. 
and Ph.D. degrees from Australian National University, 
Acton, Australia, in 2007 and 2011, respectively. 
She is currently a researcher in the Defence Science 
and Technology Group of the Australian Department of 
Defence. Her current research interests include computer 
vision and machine learning.
\end{IEEEbiography}
\vskip -1.5\baselineskip plus -1fil
\begin{IEEEbiography}[{\includegraphics[width=1in,
height=1.25in,clip,keepaspectratio]{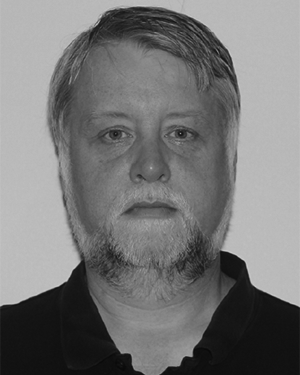}}]
{Paul Montague} received the PhD degree in mathematical 
physics from the University of Cambridge. Following 
a research career in mathematical physics, he 
transitioned into computer security and developed 
security and cryptographic solutions for Motorola. 
He is currently a researcher in computer security 
in the Defence Science and Technology Group of the 
Australian Department of Defence. His research interests 
include the application of machine learning and data 
analytic techniques to cyber security problems, and 
the robustness of machine learning.
\end{IEEEbiography}

\vfill

\EOD

\end{document}